# COMPUTATION OF VARIANCES IN CAUSAL NETWORKS

Richard E. Neapolitan
Computer Science Department
Northeastern Illinois University
Chicago, Illinois 60625

James R. Kenevan
Computer Science Department
Illinois Institute of Technology
Chicago, Illinois 60616

## ABSTRACT

The causal (belief) network is a well–known graphical structure for representing independencies in a joint probability distribution. The exact methods and the approximation methods, which perform probabilistic inference in causal networks, often treat the conditional probabilities which are stored in the network as certain values. However, if one takes either a subjectivistic or a limiting frequency approach to probability, one can never be certain of probability values. An algorithm for probabilistic inference should not only be capable of reporting the inferred probabilities; it should also be capable of reporting the uncertainty in these probabilities relative to the uncertainty in the probabilities which are stored in the network. In section 2 of this paper a method is given for determining the prior variances of the probabilities of all the nodes. Section 3 contains an approximation method for determining the variances in inferred probabilities.

## 1. INTRODUCTION

Much recent research in decision analysis and expert systems has focused on causal (belief) networks. A causal network consists of a DAG=(V,E) in which each $v \epsilon V$ represents a set of mutually exclusive and exhaustive events, along with a joint probability distribution on the alternatives of the nodes in V. The structure of the DAG represents independencies in the probability distribution. (See [Neapolitan, 1990a] for a detailed discussion of causal networks.) Two important problems in causal networks are probability propagation and abductive inference. Probability propagation is the determination of the conditional (inferred) probabilities of all nodes in the network given that evidence is obtained for the values of certain nodes, while abductive inference is the determination of the most probable, second most probable, and so on values of a specified set of nodes called the explanation set given that evidence is obtained. Pearl [1986] and Lauritzen and Spiegelhalter [1988] have obtained efficient algorithms for probability propagation for certain classes of networks, while Cooper [1984] and Peng and Reggia [1987] have obtained algorithms which perform abductive inference for certain classes of networks. The development of efficient general purpose algorithms for probability propagation and abductive inference appears unlikely since Cooper [1988] has shown that both these problems are NP–hard. Recent research has therefore centered on the development of approximation, special case, and heuristic methods. (See [Neapolitan, 1990a] for a summary of these methods.) An important recent method, developed by Chavez and Cooper [1990], uses stochastic simulation and is able to give a priori bounds for its running time as a function of relative or interval error.

The above methods treat the conditional probabilities which are stored in the network as certain values. However, if one takes either a subjectivistic or a limiting frequency approach to probability, one can never be certain of probability values. Only a pure logical approach claims to know probabilities for certain. An exact algorithm for probability propagation should not only be capable of reporting the inferred probabilities; it should also be capable of reporting the uncertainty in these probabilities relative to the uncertainty in the probabilities which are stored in the network. An approximation algorithm should incorporate this uncertainty into the possible error which is reported for the approximating values. Recent experience of one of these authors illustrates the importance of being able to report the uncertainty in probability values in medical applications: "When I was considering cervical disk surgery, I was not interested in reaching a decision by considering a 'gamble' as is done when a causal network includes decision nodes and a value node. (Such a causal network is called an influence diagram. See Shachter [1988].) Rather I only wanted to know the probability of significant improvement and the probability of a negative outcome. When a neurologist informed me that it was a fairly safe procedure, I was comforted somewhat. However, the safety of the procedure remained somewhat of a haze to me. When I learned of a study of cervical disk surgeries in which there was a 1.5% rate of minor complications and no deaths in 736 patients, I was far more confident in the safety of the procedure. I would have been less confident had the study included 73 patients."

Results of Zabell [1982] show that in many of the situations involving repeatable experiments the uncertainty in probability values must be represented by Dirichlet distributions. Neapolitan [1990a] notes that these situations often occur in expert systems applications. Using a method developed by Spiegelhalter [1988], Neapolitan [1990a] shows how to 'discretize' the Dirichlet distributions and represent the uncertainty in the probabilities which are stored in the network in the natural framework of the causal network. For example, in Figure 1 the node C represents the uncertainty in the prior probability of A while D and E represent the uncertainty in the conditional probability of B given A. Neapolitan [1990a] further shows how to use one of the algorithms for exact probability propagation to compute the variance in an inferred probability relative to the uncertainty in the probabilities which are stored in the network. The variance is clearly a good measure of the uncertainty in an inferred probability. Neapolitan notes, however, that the number of calculations needed in this computation can grow exponentially with the distance in the graph of a given node from the instantiated node. This is true even in sparsely connected networks for which exact probability propagation is feasible.

Thus there is a need for a method for determining the variances in the inferred probabilities of all nodes relative to the uncertainty in the stored probabilities. In section 2 a method is given for determining the prior variances of the probabilities of all nodes. A method for determining the variances of inferred probabilities appears very difficult even in the case of sparsely connected networks. An approximation method for determining these variances, which works only for networks in which exact probability propagation is possible, is given in section 3.

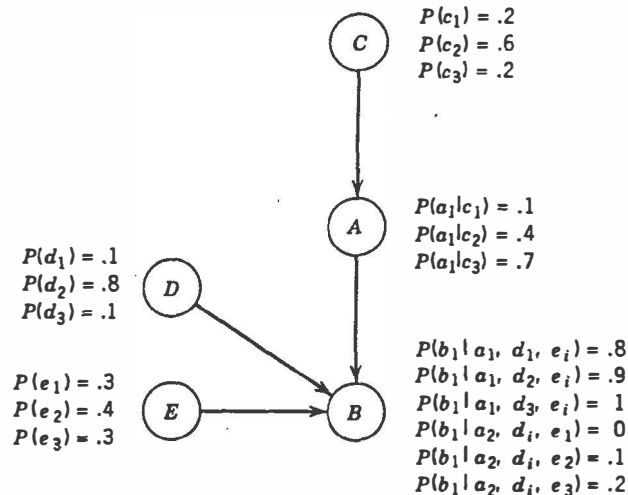

FIGURE 1. The nodes C, D, and E represent the uncertainty in probabilities.

## 2. A METHOD FOR DETERMINING PRIOR VARIANCES

It is assumed in what follows that probabilistic assessments in the causal network are made independently. Thus the uncertainties in the assessed probabilities can be represented by a set of mutually independent auxiliary parent nodes. The auxiliary parent of a node, E, will be denoted $U_E$. For example, in Figure 2, $U_E$ represents the uncertainty in $P(E)$, the prior probability of E, $U_F$ represents the uncertainty in $P(F|E)$, the conditional probability of F given E, and $U_D$ represents the uncertainty in $P(D|F,C)$. Each auxiliary node is actually a set of mutually independent nodes, one for each combination of values of the true parents. For example, $U_E$ consists of one node, if E has three alternatives, $U_F$ consists of three nodes, and if F and C each have two alternatives, $U_D$ consists of four nodes. If $U$ denotes the set of all the uncertainty nodes, then the underlying distribution is the joint probability distribution, $P(U)$, on the members of $U$. The probabilistic assessments are random variables on this joint probability distribution. Small p will be used to denote these random variables. For example, $p(e_i)$ is the random variable for $P(e_i | U)$, the prior probability of $e_i$ given values of the uncertainty variables. Similarly, $p(f_i | e_j)$ is the random variable for $P(f_i | e_j, U)$. It is assumed, for example, that $p(e_1)$ is a function only of $U_E$, $p(f_1 | e_1)$ is a function only of the first member of $U_F$, and $p(f_1 | e_2)$ is a function only of the second member of $U_F$. Therefore these random variables are mutually independent. Note however that $p(e_i)$ and $p(e_k)$ are not independent. For example, if E has two alternatives and $p(e_1) = .4$ then $p(e_2)$ must equal .6.

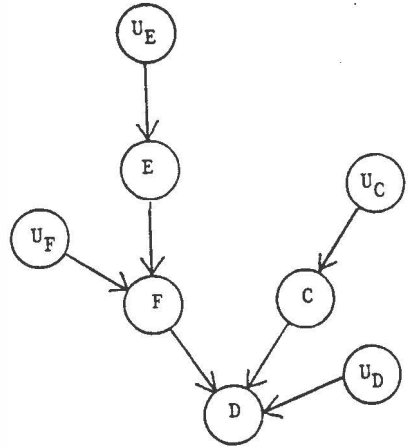

**FIGURE 2.** The auxiliary parent nodes represent the uncertainty in probabilities.

The random variable for the probability of a node which is not a root is computed from the assessed random variables. For example,

$$p(f_1) = \sum_i p(f_1 | e_i) p(e_i)$$

$$p(d_1) = \sum_{i,k} p(d_1 | f_i, c_k) p(f_i, c_k) = \sum_{i,k} p(d_1 | f_i, c_k) p(f_i) p(c_k)$$

since $p(f_i)$ and $p(c_k)$ are independent random variables.

If $p_i$ is a random variable for a probability value which is stored in the network (e.g. $p_i$ may be $p(e_i)$ or $p(f_i | e_j)$), it is assumed in this paper that the following information is available for $p_i$:

$$E(p_i) \qquad E(p_i^2) \qquad E(p_i p_j), \tag{1}$$

where E stands for expected value. The formulas for this information will be given here in the case where the distributions are Dirichlet. (See [Neapolitan, 1990a] for a detailed discussion of the Dirichlet distribution.). The Dirichlet distribution for t mutually exclusive and exhaustive events is specified by t numbers, $a_i$. For a given Dirichlet distribution, Neapolitan [1990b] shows that the information in (1) is given by

$$E(p_i) = \frac{a_i + 1}{\sum_k a_k + t} \qquad E(p_i^2) = \frac{a_i + 2}{\sum_k a_k + t + 1} \times E(p_i)$$

$$E(p_i p_j) = \frac{(a_i + 1)(a_j + 1)}{\left[\sum_k a_k + t + 1\right]\left[\sum_k a_k + t\right]}.$$

## 2.1. The Case of Trees

Since the variance, $V(p)$, is given by $E(p_i^2) - (E(p_i))^2$, the prior variances of all roots can be computed from the information listed in (1) for the probabilities of the roots. Next we will consider the case of computing the prior variance in a node, F, with one parent from the information listed in (1) for the probability of the parent, E, along with the information listed in (1) for the conditional probabilities of the node given values of its parent. Then the information listed in (1) (and therefore the variances) can be computed for all nodes using a downward propagation scheme. We have first that

$$p(f_i) = \sum_k p(f_i|e_k)p(e_k) \quad \text{and thus} \quad E(p(f_i)) = \sum_k E(p(f_i|e_k))E(p(e_k)).$$

The latter expression can be computed from the information listed in (1) for $p(E)$ and $p(F|E)$. Furthermore

$$E(p(f_i)p(f_j)) = E(\sum_k p(f_i|e_k)p(e_k) \times \sum_k p(f_j|e_k)p(e_k)).$$

This expression is the sum of terms which have the following form:

$$E(p(f_i|e_k)p(f_j|e_k)p(e_k)^2) \quad \text{and} \quad E(p(f_i|e_k)p(e_k)p(f_j|e_m)p(e_m)).$$

Now
$$E(p(f_i|e_k)p(f_j|e_k)p(e_k)^2) = E(p(f_i|e_k)p(f_j|e_k)) \times E(p(e_k)^2)$$

$$E(p(f_i|e_k)p(e_k)p(f_j|e_m)p(e_m)) = E(p(f_i|e_k)) \times E(p(f_j|e_m)) \times E(p(e_k)p(e_m)).$$

Thus these terms can be computed from the information listed in (1) for $p(E)$ and $p(F|E)$.

Finally
$$E(p(f_i)^2) = E((\sum_{k=1}^t p(f_i|e_k)p(e_k))^2).$$

This expression is the sum of the following types of terms:

$$E(p(f_i|e_k)^2)E(p(e_k)^2) \quad \text{and} \quad E(p(f_i|e_k))E(p(f_i|e_m))E(p(e_k)p(e_m)).$$

These terms can also be computed from the information listed in (1) for $p(E)$ and $p(F|E)$.

It is clear that the dominant computations in the above algorithm are the ones that determine $E(p(f_i)p(f_j))$. If T is the maximum number of alternatives for a node, then the maximal number of calculations in one of these determinations is easily seen to be $\theta(T^2)$ and the number of computations in the entire algorithm is $O(nT^4)$, where n is the number of nodes in the network.

## 2.2. Extension to Singly Connected Networks

The method outlined above can be generalized to the case where a node has more than one parent. For the sake of simplicity the generalization will be illustrated by considering the case of two parents, D and E, of a node F. Since the network is singly connected, D and E are independent and therefore

$$p(f_i) = \sum_{j,k} p(f_i|d_j,e_k)p(d_j)p(e_k) \quad \text{and thus} \quad E(p(f_i)^2) = E((\sum_{j,k} p(f_i|d_j,e_k)p(d_j)p(e_k))^2),$$

which is the sum of the following types of terms:

$$E(p(f_i|d_j,e_k)^2)E(p(d_j)^2)E(p(e_k)^2)$$
$$E(p(f_i|d_j,e_k))E(p(f_i|d_j,e_m))E(p(d_j)^2)E(p(e_k)p(e_m))$$
$$E(p(f_i|d_j,e_k))E(p(f_i|d_n,e_k))E(p(d_j)(p(d_n))E(p(e_k)^2)$$
$$E(p(f_i|d_j,e_k))E(p(f_i|d_n,e_m))E(p(d_j)p(d_n))E(p(e_k)p(e_m)).$$

All of these terms can be computed from the information listed in (1) for $p(D)$, $p(E)$, and $p(F|D,E)$. Similar results can readily be obtained for $E(p(f_i))$ and $E(p(f_i)p(f_j))$. Thus a downward propagation algorithm can be developed as in section 2.1.

As in the case of trees, the dominant computations are the ones that determine $E(p(f_i)p(f_j))$. If T is the maximum number of alternatives for a node, m is the minimal number of parents for a node, and M is the maximal number of parents for a node, then the number of calculations in the above algorithm is $\Omega(T^{2m+2})$ and the number of computations in the entire algorithm is $O(nT^{2M+2})$, where n is the number of nodes in the network. Furthermore, if v is the number of specified values in the network then the number of computations in the algorithm is $O(v^2)$.

Example. Suppose that we have a tree with two nodes, E and F, each with two alternatives, and that there is an arc from E to F. Suppose further that prior ignorance in $P(e_1)$, $P(f_1|e_1)$, and $P(f_1|e_2)$ is represented by using the symmetric Dirichlet distribution in which $a_1=a_2=0$ for all these probabilities. If $p_1$ stands for $p(e_1)$, $p(f_1|e_1)$, or $p(f_1|e_2)$, and $p_2$ stands respectively for $p(e_2)$, $p(f_2|e_1)$, or $p(f_2|e_2)$, then using the formulas given above for obtaining the information in (1) in the case of the Dirichlet distribution, we have that

$$E(p_i) = 1/2 \qquad E(p_i^2) = 1/3 \qquad E(p_1 p_2) = 1/6.$$

Thus
$$V(p_i) = E(p_i^2) - (E(p_i))^2 = ((1/3)^2 - (1/2)^2 = .0833$$

$$E(p(f_1)) = E(p(f_1|e_1))E(p(e_1)) + E(p(f_1|e_2))E(p(e_2)) = (1/2)(1/2) + (1/2)(1/2) = .5$$

$$E(p(f_1)^2) = E(p(f_1|e_1)^2)E(p(e_1)^2) + 2(E(p(f_1|e_1)p(f_1|e_2))E(p(e_1)p(e_2))$$
$$+ E(p(f_1|e_2)^2)E(p(e_2)^2)$$
$$= (1/3)(1/3) + 2(1/2)(1/2)(1/6) + (1/3)(1/3) = .3056$$

$$V(p(f_1)) = E(p(f_1)^2) - (E(p(f_1)))^2 = .3056 - .25 = .0556.$$

It may seem a bit odd that one is more certain in the probability of F than in any of the probabilities which are stored in the network. An intuitive basis for this result can be obtained by considering the following situation. Suppose that three urns, U, V, and W, each contains two coins, and in each of the urns the 'propensity' for the first coin landing heads in a random toss has an associated probability of .25 and the 'propensity' for the second coin landing heads has an associated probability of .75. Suppose further that we choose a coin at random from each urn, and toss it. If heads comes up on the coin from urn U then we inspect the coin chosen from urn V, while if tails comes up on the coin from urn U then we inspect the coin chosen from urn W. The three urns can be likened to the uncertainty variables in the above example. Let $F$ be the event that the coin we inspect shows a head. It is easy to see that the prior probability (before any coins are chosen) of any of the coins showing a head is .5 and the prior probability of $F$ is .5. Suppose next that we consider the space consisting of the 8 possible combinations of coins, and consider the probabilities of heads on each of the coins and the probability of $F$ as random variables in this space. Clearly the probability of heads on each of the coins can obtain values .25 and .75 with equal probability. It is easy to show, however, that the probability of $F$ can obtain values .25, .375, .625, and .75 with equal probability. Thus if we choose some combination of coins and toss that combination many times, the fraction of times $F$ occurs could fall closer to .5 than the fraction of heads on any one of the coins. If we consider the 8 combinations of coins as 'possible worlds' then it is possible that the probability of $F$ is close to .5 (i.e. it could be .375 or .625) while it is not possible that the probability of heads on any one of the tosses is close to .5.

### 2.3. A Downward Propagation Scheme for Determining the Variances in Inferred Probabilities

When evidence is obtained, the variances in the inferred probabilities of the nodes will in general be different from their prior variances. Only a downward propagation scheme has been developed. The development of an exact upward propagation method appears very difficult. This problem will be addressed in section 3. Consider the tree in Figure 3, and suppose that D is instantiated for $d_1$. We wish to determine the variances in the inferred probabilities of E and F. The probability distribution for $p(e_j|d_1)$ is one of the distributions which is stored in the the network. Thus E can be considered a 'root' with this probability distribution. We have that

$$p(f_i|d_1) = \sum_j p(f_i|e_j,d_1)p(e_j|d_1) = \sum_j p(f_i|e_j)p(e_j|d_1)$$

since E d–separates D from F. (See [Pearl, 1988] for a definition of d–separation.) Thus $V(p(f_i|d_1))$ can be computed using the algorithm described above where the distribution of $p(e_j|d_1)$ is used as the prior distribution for node E. Clearly, the propagation algorithm can be used for F's descendents and the scheme also applies to an arbitrary singly connected network.

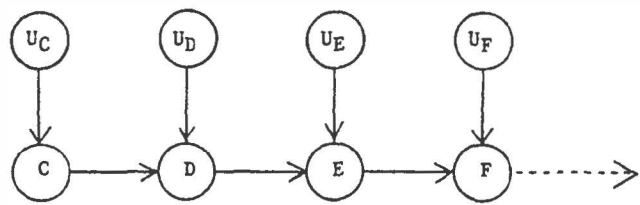

**FIGURE 3.** D is instantiated for $d_1$.

### 2.4. Extension to Nonsingly Connected Networks

The prior variances in nonsingly connected causal networks can be computed using a technique similar to Pearl's [1988] method of 'conditioning'. The method will be illustrated on the network in Figure 4. For the network in that figure we have that

$$p(f_i) = \sum_{k,j} p(f_i|d_k,e_j)p(d_k,e_j).$$

Since D and E are not independent the method described in subsection 2.2 is not immediately applicable. However, conditioning on C, we have that

$$p(f_i) = \sum_m p(f_i|c_m)p(c_m).$$

Now $\quad p(f_i|c_m) = \sum_{j,k} p(f_i|d_j,e_k,c_m)p(d_j,e_k|c_m) = \sum_{j,k} p(f_i|d_j,e_k)p(d_j|c_m)p(e_k|c_m).$

Therefore the information listed in (1) for $p(f_i|c_m)$ can be computed by instantiating C for $c_m$, considering D and E as 'roots', and using the method described in the previous subsection. Furthermore, due to the arguments in subsection 2.1, the information listed in (1) for $p(f_i)$ can be computed from the information listed in (1) for $p(f_i|c_m)$ and $p(c_m)$. Thus the only additional information needed is the information listed in (1) for node C. In this case this information is immediately available since C is root. In general this information for the nodes in the loop cutset can be computed using the chain rule and the method described in the previous subsection.

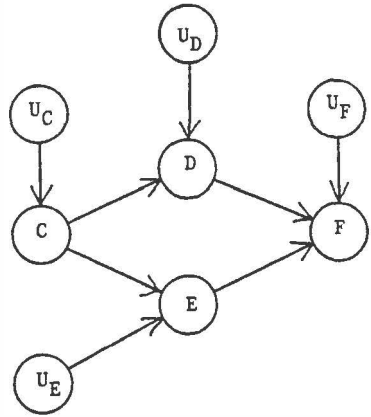

FIGURE 4. A nonsingly connected causal network.

### 3. DETERMINING THE VARIANCES IN INFERRED PROBABILITIES

This section outlines a method for determining confidence intervals for the standard deviations of the random variables which represent inferred probabilities and tolerance intervals for the distributions of those random variables. The method determines the number of computations needed to achieve an a priori specified width in the confidence intervals. The method has only been developed for networks in which exact probability propagation is possible. In order to use the method in this section one must be able to obtain the posterior distribution on the uncertainty variables. If W is the evidence, one way to this is to assume that the posterior distribution on the uncertainty variables, $P(U|W)$, can be approximated by the prior distribution, $P(U)$. Neapolitan [1990b] shows that the posterior distribution on the uncertainty variables can differ from the original distributions by no more than a distribution based on the information in one additional trial. Therefore if there is a reasonable amount of confidence in the original distributions, the posterior distributions can be approximated by the original distributions. The use of this approximation means that we are assuming that the underlying joint distribution on the auxiliary parent variables remains unchanged when variables are instantiated. Therefore these variables remain independent. Note that in general roots do not remain independent when variables are instantiated. In the case where there is not a reasonable amount of confidence in the stored probabilities techniques from [Spiegelhalter and Lauritzen, 1990] can be used to approximate the posterior distributions. However, the method outlined here is only appropriate if it can be assumed that the auxiliary parent variables remain independent.

Due to the assumption that the auxiliary parent variables remain independent, a sample point from the entire space can be obtained by independently randomly generating a value for each probability which is stored in the network. The method for randomly generating the probability values at a node will be illustrated with a root, E, which has 3 alternatives. Suppose that $\mu_1(p)$ is a Dirichlet distribution for the random variable $p(e_1)$. Randomly generate a number r in the interval [0,1]. Then solve the following equation for x:

$$\int_0^x \mu_1(p)dp = r.$$

The randomly generated value of $p(e_1)$ is x. Next determine the conditional distribution of $p(e_2)$ given that $p(e_1)=x$. It can be shown that this distribution is the following Dirichlet distribution:

$$\mu_2(p) = \frac{(a_2+a_3+1)!}{a_2!a_3!} p^{a_2}(1-p)^{a_3}dp$$

where $p=p(e_2)/(1-x)$. Repeat the above procedure to obtain a random value, y, of p. The randomly generated value of $p(e_2)$ is then $(1-x)y$. Finally the randomly generated value of $p(e_3)$ is one minus the sum of the other two randomly generated values.

The set of all randomly generated probability values uniquely determines a causal network in which there is no uncertainty in the probability values. Probability propagation can then be performed in this network using one of the algorithms for exact propagation. The computed probability of a value of a node represents one sample point for the random variable which represents the probability of that value. The expected value of the random variable can be obtained by performing exact propagation using the expected values of the Dirichlet distributions. A confidence interval for the standard deviation in the random variable can then be obtained using the following standard method. (See [Hogg and Craig, 1970].) Suppose a 95% confidence interval is desired. Let

$$a(n) = \tfrac{1}{2}(z_{.975} + \sqrt{2n-1})^2 \quad \text{and} \quad b(n) = \tfrac{1}{2}(z_{.025} + \sqrt{2n-1})^2$$

where the values of $z_Q$ can be obtained from the bottom row of a table for the chi–square distribution and n is the number of trials. In particular

$$z_{.975} = -1.96 \quad \text{and} \quad z_{.025} = 1.96.$$

Then if n>100 the following is a 95% confidence interval for the standard deviation:

$$\left[\left[\frac{\sum_i (x_i - E)^2}{b(n)}\right]^{1/2}, \left[\frac{\sum_i (x_i - E)^2}{a(n)}\right]^{1/2}\right]$$

where E is the expected value of the random variable and $x_i$ is the value on the ith trial. It is easy to show that if we want the confidence interval to be smaller than $\epsilon$ then it is necessary that

$$\left[\sum_i (x_i-E)^2\right]^{1/2} \times \left[\frac{b(n)^{1/2}-a(n)^{1/2}}{a(n)^{1/2}b(n)^{1/2}}\right] < \epsilon.$$

Since $0 \leq x_i \leq 1$ it is sufficient to choose n so that

$$n^{1/2} \times \text{MAX}\left[E,(1-E)\right] \times \left[\frac{b(n)^{1/2}-a(n)^{1/2}}{a(n)^{1/2}b(n)^{1/2}}\right] < \epsilon.$$

The value of n can be computed numerically. For example, if $\epsilon=.1$ and $E=.5$ it is sufficient to take 200 trials to obtain a 95% confidence interval. Notice that we have used extreme values of 0 and 1 for $x_i$ to arrive at this n. In general, the $x_i$ will often fall much closer to E and therefore the width of the confidence interval obtained will be much smaller than the specified value of $\epsilon$. Furthermore, one should be able to improve on the accuracy of the results by using an equidistributed sequence instead of random or pseudorandom numbers. (See [Davis and Rabinowitz, 1975] for a discussion of equidistributed sequences.)

Also of interest would be a value of n which would make the ratio of the width of the confidence interval to E smaller than a specified $\epsilon$. It is easy to see that it is sufficient to choose n so that

$$n^{1/2} \times \text{MAX}\left[1, \frac{1-E}{E}\right] \times \left[\frac{b(n)^{1/2} - a(n)^{1/2}}{a(n)^{1/2} b(n)^{1/2}}\right] < \epsilon.$$

Note that the sufficient value of n approaches $\infty$ as E approaches 0. Chavez and Cooper [1990] discuss a way to handle expected probabilities which are equal to 0. However, in the case where all the distributions are Dirichlet, no expected probabilities will be equal to 0.

Finally one would also be interested in obtaining a tolerance interval for the distribution of a random variable which represents an inferred probability. A tolerance interval is an interval such that, given specified p and $\gamma$ where $0 \leq p, \gamma \leq 1$, the probability that the interval contains at least 100p per cent of the distribution is equal to $\gamma$. Using nonparametric methods (See [Hogg and Craig, 1970].) such an interval can be obtained in the following way. Suppose the sample has size n, u is the smallest value in the sample and U is the largest value. Given a specified p, if we let

$$\gamma = 1 + (n-1)p^n - np^{(n-1)}$$

then nonparametric methods yield [u,U] as a 100$\gamma$ per cent tolerance interval for 100p per cent of the distribution. If $\gamma$ and p are specified, the above equation can be solved numerically for n to yield a sufficient value of n. Once the sample is obtained any 2 points from the sample can be chosen to obtain a narrower tolerance interval. If the points are ordered $x_1 < x_2 < \cdots < x_n$, p is specified, and we set

$$\gamma' = 1 - \frac{n!}{(j-i-1)!(n-j+i)!} \int_0^p x^{j-i-1}(1-x)^{n-j+i} dx,$$

then $[x_i, x_j]$ is a 100$\gamma'$ per cent tolerance interval fo 100p per cent of the distribution. Note that the distribution above is the Dirichlet distribution with $a_1 = j-i-1$ and $a_2 = n-j+i$.

The methods outlined in this section only pertain to causal networks in which exact probability propagation is feasible. There remains the problem of applying these methods to causal networks in which exact propagation is not feasible. Perhaps these methods can be combined with those of [Chavez and Cooper, 1990].

Neapolitan [1990b] has obtained an exact method for determining the variances in inferred probabilities, based on Pearl's [1986] belief propagation method, in the case where the posterior distribution on the uncertainty variables can be approximated by the prior distribution.

## 4. AN UPPER BOUND ON THE VARIANCE

If p is a random variable which represents a probability value, then $0 \leq p \leq 1$ and therefore

$$V(p) = E(p^2) - (E(p))^2 \leq E(p) - (E(p))^2.$$

203

This bound is clearly meaningless if E(p)=.5 since it yields a value of .25 in that case. (Note that .25 is the absolute maximum variance a probability value could have.) However the bound can be quite useful for large values of E(p). The above inequality implies that the ratio of the standard deviation of p to E(p) is bounded above by

$$(1/E(p) - 1)^{1/2}.$$

If E(p) is very large this bound is very small. Thus, for example, if a system determines that the expected probability of a disease being present is large, one can be highly confident that this expected probability is close to the 'correct' probability. This is true regardless of the confidence in the probabilities which are stored in the network.